\documentclass[runningheads, hidelinks]{llncs}
\usepackage{comment}
\usepackage{algorithm} 
\usepackage{algpseudocode} 
\usepackage{booktabs}
\usepackage{multirow}
\usepackage{tabularx}
\usepackage{graphicx}
\usepackage{relsize}
\usepackage{booktabs} % To thicken table lines
\usepackage{pdfpages}
\usepackage[
textwidth=0.32\textwidth,textsize=footnotesize]{todonotes}
\usepackage{enumitem}
\usepackage{ragged2e,array,url,graphicx,csquotes,hyperref,adjustbox,longtable}
\usepackage[english]{babel}
\graphicspath{ {pictures/} }
\usepackage{setspace}

\definecolor{PrologPredicate}{RGB}{0,0,200}
\definecolor{PrologVar}      {RGB}{145,032,039}
\definecolor{PrologComment}  {RGB}{169,082,044}
\definecolor{PrologOther}    {rgb}{0.2,0.2,0.2}
\definecolor{PrologString}   {rgb}{0.2,0.2,0.2}
\usepackage{listings} 
\usepackage{textcomp}
\newcommand{\code}{\lstinline[style=MyInline]}
\lstnewenvironment{scasp}[1][]{\lstset{style=MySCASP,#1}}{}
\lstdefinestyle{MyInline}
{
  basicstyle = \ttfamily\color{PrologVar},
  breaklines = true,
  breakatwhitespace=true,
  upquote = true,
  literate =
  {,}{}{1\discretionary{,}{}{,}}
  {\ │}{{$\mid$}}1 
  {|}{{$\mid$}}1
  {\\\{}{{\{}}1
  {\\\}}{{\}}}1
  {[}{{\small[}}1
  {]}{{\small]}}1
  {.=.}{{\#=}}3
  {.<.}{{\#<}}3
  {.>.}{{\#>}}3
  {.=<.}{{\#=<}}4
  {.>=.}{{\#>=}}4
  {\\=}{{\char"5C=}}2
  {?-}{{?-}}2
  {:-}{{:-}}2
  {\\$}{{\$}}1
}
\lstdefinestyle{tree}
{
  basicstyle = \scriptsize\ttfamily\color{PrologPredicate},
  basewidth = 0.46em,
  moredelim = {[s][\color{PrologString}]{ \{}{\} }},
  moredelim = {*[s][{\color{PrologVar}}]{(}{)}},
  literate     =
  {.\\=.}{{\ \char"5C=\ }}3
  {\\=}{{\ \char"5C=\ }}3
  {.<.}{{\ \#<\ }}4
  {.>.}{{\ \#>\ }}4
  {.=.}{{\ \#=\ }}3
  {.=<.}{{\ \#=<\ }}5
  {.>=.}{{\ \#>=\ }}5
}
\lstdefinestyle{Plain}
{
  keywords = {},
  upquote = true,
  basicstyle = \relsize{-0.5}\ttfamily\color{PrologPredicate},
  basewidth = 0.48em,
  numbers=none,
  xleftmargin=0cm,
  moredelim = {*[s][\color{black!40!PrologPredicate}]{\#pred}{.}},
  moredelim = {*[s][\color{black!40!PrologPredicate}]{\#show}{.}},
  moredelim = {*[s][\color{black!40!PrologPredicate}]{\#hide}{.}},
  moredelim = {*[s][\color{PrologVar}]{(}{)}},
  moredelim = {*[s][\color{PrologString}]{'}{'}},
 commentstyle = \mdseries\color{PrologComment},
  morecomment=[l]\%,
   literate     =
  {|}{{$\mid$}}1
  {\\$}{{\$}}1
  {\ │}{{$\mid$}}1,
}
\lstdefinestyle{MySCASP}
{
  keywords = {},
  basicstyle = \relsize{-0.5}\ttfamily\color{PrologPredicate},
  basewidth = 0.48em,
 moredelim = {**[is][\color{PrologOther}]{`}{`}},
  moredelim = {*[s][\color{PrologOther}]{:-}{.}},
  moredelim = {*[s][\color{black!60!PrologPredicate}]{\#pred}{.}},
  moredelim = {*[s][\color{black!60!PrologPredicate}]{\#show}{.}},
  moredelim = {*[s][\color{black!60!PrologPredicate}]{\#hide}{.}},
  moredelim = {*[s][\color{PrologVar}]{(}{)}},
  commentstyle = \mdseries\color{PrologComment},
  morecomment=[l]\%,
  literate     =
  {&(}{{\color{PrologOther}(}}1
  {&)}{{\color{PrologOther})}}1
}
\newcommand{\blist}{\smallskip\begin{list}{$\bullet$}{\topsep=1pt \parsep=0pt \itemsep=4pt}}
\newcommand{\elist}{\end{list}\medskip}

\newcommand{\bnum}{\begin{list}{}{\topsep=2pt \parsep=0pt \itemsep=1pt}}
\newcommand{\enum}{\end{list}\medskip}

\begin{document}

\title{Counterfactual Explanation Generation with s(CASP)}%\thanks{%
  % We thank the anonymous reviewers for suggestions for improvement. %
 % }}
 
%\begin{comment}
\author{%
  Sopam Dasgupta$^1$\orcidID{0009-0008-3594-5430}, \quad
  Farhad Shakerin$^2$, \quad
  Joaqu\'in Arias$^3$\orcidID{0000-0003-4148-311X}, \quad
  Elmer Salazar$^1$, \quad
  Gopal Gupta$^1$\orcidID{0000-0001-9727-0362} \\
  \institute{$^{1}$The University of Texas at Dallas, Richardson, USA \\ 
  $^2$ {Microsoft}\\
    $^3$CETINIA, Universidad Rey Juan Carlos, Madrid, Spain}
  \email{\{sopam.dasgupta, ees101020, gupta\}@utdallas.edu} \quad \quad
  \email{fshakerin@microsoft.com} \quad \quad
  \email{joaquin.arias@urjc.es}
}
\authorrunning{Dasgupta et al.}   
%\end{comment}

%
\maketitle              % typeset the header of the contribution
%

% \vspace{-1em}

%\todo[inline]{JAH: The review process of PADL 2024 is double-anonymous. In your submission, please, omit your names and institutions; refer to your prior work in the third person, just as you refer to prior work by others; do not include acknowledgments that might identify you.}

\begin{abstract}

Machine learning models that automate decision-making are increasingly being used in consequential areas such as loan approvals, pretrial bail, hiring, and many more. 
%Due to the rise of extremely powerful and accurate predictive models, they are increasingly used in these consequential decision-making processes, directly or indirectly. 
Unfortunately, most of these models are black-boxes, i.e., they are unable to reveal how they reach these prediction decisions. A need for transparency demands justification for such predictions. An affected individual might desire explanations to understand why a decision was made. Ethical and legal considerations may further require informing the individual of changes in the input attribute that could be made to produce a desirable outcome. This paper focuses on the latter problem of automatically generating   \textit{counterfactual explanations}. Our approach utilizes answer set programming and the s(CASP) goal-directed ASP system. Answer Set Programming (ASP) is a well-known knowledge representation and reasoning paradigm. s(CASP) is a goal-directed ASP system that executes answer-set programs top-down without grounding them. The query-driven nature of s(CASP) allows us to provide justifications as proof trees, which makes it possible to analyze the generated counterfactual explanations. We show how counterfactual explanations are computed and justified by imagining multiple possible worlds where some or all factual assumptions are untrue and, more importantly, how we can navigate between these worlds. We also show how our algorithm can be used to find the \textit{Craig Interpolant} for a class of answer set programs for a failing query.

\end{abstract}

\section{Introduction}
Predictive models are used in decision-making, for example, in the filtering process for hiring candidates for a job or approving a loan. Unfortunately, most of these models are like a black box, making understanding the reasoning behind a decision difficult. In addition, many of the decisions such models make have consequences for humans affected by them. Humans, subject to unfavorable decisions/judgments/outcomes, desire a satisfactory explanation. This desire for transparency is essential, regardless of whether a system (e.g., a data-driven prediction model) or humans make such consequential decisions. Hence, there is a challenge to make such consequential decisions explainable to humans. Wachter et al. \cite{wachter} highlighted a possible explanation for the decision made in the form of a counterfactual that explains the reasoning behind a decision and informs a user on how to achieve the desired outcome. Our efforts in this paper focus on achieving the desired outcomes, given that one has received a negative decision from a macine-learned model. We attempt to answer the question, `What can be done to achieve the desired outcome?' Our approach models various worlds. The idea is to move from the current world where the decision outcome is unfavorable to a world(s) where the desired outcome would hold given the static decision-making process (i.e., the decision maker does not change). The traversal between these worlds, the original world where we get the undesired outcome, to the counterfactual world(s) where we get the desired decision, is done through \textit{interventions}, i.e., by changing the input \textit{feature values}.

\section{Background}

\vspace{-0.1in}
\subsection{Counterfactual Reasoning}
As humans, we treat explanations as tools to help us understand decisions and inform us on how to act. Wachter et al. \cite{wachter} argued that for individual decisions, \textit{counterfactual explanations (CFE)} should be used as a means to provide explanations. \textit{Counterfactual explanations (CFE)} offer meaningful explanations to understand a decision and inform on what can be done to change the outcome to a desired one. In the context of being denied a loan, \textit{counterfactual explanations (CFE)} are similar to the statement: \textit{If your credit score was 730, you would have been approved for a loan.} A key idea behind a counterfactual is imagining a different world where the desired outcome would hold. This different world ought to be reachable from the current world. Thus, the concept of ``closest" or ``close possible worlds" imagines alternate (reasonably plausible) scenarios where such a desired outcome would be achievable.

For a prediction model given by $f:X \rightarrow \{0,1\}$, we define a set of counterfactual explanations $\hat{x}$ for a factual input $x \in X$ as $\textit{CF}_{f}(\hat{x})=\{\hat{x} \in X | f(x) \neq f(\hat{x})\}$. The set of counterfactual explanations contain all the inputs that correspond to the predictive model $f$ returning a different prediction from the original factual input $x$. % Sopam_10_5_begin 
We show how counterfactual reasoning can be performed using the s(CASP) query-driven predicate ASP system. By utilizing s(CASP)'s inbuilt \textit{dual rules}, counterfactual explanations can be naturally obtained. Given the definition of a predicate {\tt p} as a rule in ASP, its corresponding dual rule allows us to prove {\tt not p}, where {\tt not} represents negation as failure. We utilize these dual rules to construct alternate worlds then use them to obtain counterfactual explanations. We highlight this property later in the paper by showing the case of the adult dataset to check whether somebody makes `$=<$\$50k/year', or `$>$\$50k/year'. 

\vspace{-0.1in} 
\subsection{ASP, s(CASP), and Commonsense Reasoning}

Answer Set Programming (ASP) is a well established paradigm for knowledge representation and reasoning \cite{baral,gelfond-kahl,cacm-asp}. s(CASP) is a goal-directed ASP system that executes answer set programs in a top-down manner without grounding them \cite{scasp-iclp2018,ref_GG}. The query-driven nature of s(CASP) greatly facilitates performing commonsense reasoning as well as counterfactual reasoning based on our commonsense knowledge. Additionally, s(CASP) utilizes proof trees which justify counterfactual explanations

Commonsense knowledge in ASP can be emulated using (i) default rules, (ii) integrity constraints, and (iii) multiple possible worlds~\cite{ref_GG,gelfond-kahl}. Default rules are used for jumping to conclusions unless an exception applies, e.g., a bird flies unless it is a penguin. Thus, if we are told that Pingu is a bird, we conclude that Pingu flies. Later, if we are told that Pingu is a penguin, we withdraw the conclusion that Pingu can fly. Default rules with exceptions represent bulk of human knowledge~\cite{gelfond-kahl}. 
For the given example, we have two birds, Tweety and Pingu. Additionally, Pingu is a penguin. We model the knowledge for the function \code{fly}{\tt (X)} through the rule in line 2. We 
notice an exception in the form of {\tt not} \code{ab1}{\tt (X)}. This carves an exception for birds that do not fly, such as penguins.
\begin{lstlisting}[style=MySCASP, firstnumber=1]
bird(tweety).          bird(pingu).         penguin(pingu).
fly(X) :- bird(X), not ab1(X).
ab1(X) :- penguin(X).
\end{lstlisting}

\noindent 
For the given program, s(CASP) implicitly generates dual rules that formalize the conditions for negating the rules. 

Integrity constraints allow us to express impossible situations and invariants. 
For example, we know that if an object is a bird, it cannot be a mammal.
%For example, we know that the color blue is not the color red. Suppose we have a world where each object can only have one color. 
Integrity constraints help remove certain undesired worlds (those worlds where an object could be both a bird as well as a mammal, unless it's a bat) from the multiple worlds we consider to compute counterfactual explanations.

\begin{lstlisting}[style=MySCASP]
:- bird(X), mammal(X), not bat(X).
\end{lstlisting}

\noindent 
Which is essentially the same as 
\begin{lstlisting}[style=MySCASP]
false :- bird(X), mammal(X), not bat(X).
\end{lstlisting}

%\medskip Utilizing NAF(negation-as-failure), which allows us to construct multiple unique worlds, each representing alternate universes, they show possible scenarios where things might have been different. 

\begin{comment}
\noindent Negation as failure also allows us to represent multiple possible worlds. For example, in the real world, penguins cannot fly. However, we could imagine a cartoon world of super-powered animals that can fly, which may share many similarities with our natural world (e.g., normal birds can fly in both worlds). Penguins cannot fly in the former but can in the latter because we imagine them to be able to.   
\end{comment}

% Sopam_10_4_Begin
In ASP, multiple worlds can be specified through the use of even loops over negation (or even cycles) \cite{ref_GG}.
Even cycles are loops with an even number of intervening negations between a call and its recursive invocation in the dependency graph \cite{gelfond-kahl}. %When an even cycle is detected, we assume success. The goal that depends on that even cycle will always succeed. 
Even cycles determine the possible worlds that can be generated. For example, the following rules will generate two worlds, one in which John teaches databases and Mary does not, and another one in which Mary teaches databases and John does not.

\begin{lstlisting}[style=MySCASP]
teaches_db(mary) :- not teaches_db(john).
teaches_db(john) :- not teaches_db(mary). 
\end{lstlisting}

In s(CASP), the query {\tt ?- teaches\_db(X)} will produce two answers: \{{\tt{teaches\_db(mary),  not teaches\_db(john)}}\} and \{{\tt {teaches\_db(mary),  not teaches\_db(john)}}\}.
%\usepackage{multirow}
%
\begin{comment}
{\tt { teaches\_db(mary),  not teaches\_db(john) } 

X = mary ? ;

\smallskip 
{ teaches\_db(john),  not teaches\_db(mary) } 

X = john ? 
}

X = mary ? ;

\smallskip 
{ teaches\_db(john),  not teaches\_db(mary) } 

X = john ?  
\end{comment}
%
Note that s(CASP) produces partial answer set, so it lists both positive and negative atoms.
%
\iffalse 
So a predicate {\tt p} in an even cycle will be true in one world and false in another. \newline If we query {\tt ?- jill\_eats(X)}, then the predicate {\tt jill\_eats/1} will be true if {\tt jack\_eats/1} is not true. From the dual program, {\tt jack\_eats/1} will be false if {\tt jill\_eats/1} is true. So, {\tt jill\_eats/1} depends on itself and there are two intervening negations: one in rule 2 and one in rule 1. In this case, we assume jill\_eats(X) is true.
\begin{lstlisting}[style=MySCASP]
jack_eats(X):- not jill_eats(X).
jill_eats(X):- not jack_eats(X).
\end{lstlisting}
\medskip Similarly if we run the query {\tt ?- jack\_eats(X)}, it will expand into not jill\_eats(X) which in turn not (not jack\_eats(X)) where in we assume {\tt jack\_eats(X)} is true. We have two possibilities: either {\tt jill\_eats(X)} is true or {\tt jack\_eats(X)} is true. This enables the creation of multiple possible models. 
\fi 
%
Even cycles also permit abductive reasoning which  generates explanations that justify observations. \textit{Abducibles} can also be represented in s(CASP) with simple even cycles. More details on abduction can be found elsewhere \cite{ref_GG,gelfond-kahl}. 

% \todo[inline]{JAH: I would change the order a little:\\
% Section 3: Use case: the Adult Income Dataset\\
% Section 4: Find counterfactual using s(CASP)\\
% Subsection 4.1: Modeling Factual Knowledge
% Subsection ...}
% \todo[inline]{JAH: Additionally, I think there are many subsections...}

\vspace{-0.1in} 
\subsection{Craig Interpolant}

% Correct grammar
Craig's interpolation theorem \cite{ref_craig} in mathematical logic states that if $A \rightarrow C$ is a valid implication, there is a Craig Interpolant such that $A \rightarrow I$ and $I \rightarrow C$ are valid, and every non-logical symbol of $I$ exists in both $A$ and $C$. 
A reverse form of this theorem exists that states: ``For two subsets of clauses such that $A \wedge B$ is unsatisfiable, there exists a reverse interpolant $I$ such that $A \rightarrow I$ and $B \rightarrow \neg I$ and every non-logical symbol of I exist in both $A$ and $B$." This reverse form has important aspects in model checking. McMillan \cite{ref_McMillan_1,ref_McMillan_2} proposed a systemic way of computing interpolants for inconsistent sets $(A,B)$ using the resolution proof of unsatisfiability $P$ of $A \cup B$. 
Using this proof\; we can find the source of the inconsistency which on resolving makes $A \wedge B$ satisfiable.
In our case we have an inconsistency in $(A,B)$ where $A$ represents a factual instance where one gets an undesired outcome and B represents the rules that correspond to the world where one would obtain a desired result. The proof of the inconsistency or the Craig Interpolant will indicate what needs to be resolved such that $(A, B)$ is consistent

\vspace{-0.1in}
\subsection{FOLD-SE}
FOLD-SE \cite{foldse} is an efficient, explainable, rule-based machine learning algorithm for classification tasks. It comes from the FOLD (First Order Learner of Defaults) family of algorithms  \cite{fold,foldrpp,foldrm}. For given input data (numerical and categorical), FOLD-SE generates a set of default rules—essentially a stratified normal logic program—as an (explainable) trained model. The explainability obtained through FOLD-SE is scalable. Regardless of the size of the dataset, the number of learned rules and learned literals stay relatively small while retaining good accuracy in classification. It is comparable with the likes of XGBoost \cite{xgboost} and Multi-Layer Perceptrons (MLP) with the added advantage of being explainable.

\vspace{-0.1in}
\subsection{Causality and Causal Relations}
Implicit assumptions are made where changes resulting from  interventions will be independent of changes across features. However, this is only true in worlds where features are independent. This assumption of independence across features might need to be revised in the real world. Causal Relationships that govern the world should be taken into account. For example, Consider the case where an individual wants to be approved for a loan. Given that the loan approval system takes into account the outstanding dues, the savings of the individual, and the credit score, it might make a recommendation of raising the credit score by 30 points to 700 for an individual with savings(X$_1$) = \$25,000, outstanding dues(X$_2$) = \$10,000 and credit score(X$_3$) = 670. However, the credit score depends on many factors, including the outstanding dues. Hence, a functional dependency exists between at least two features (outstanding dues and credit score). In addition, clearing the outstanding dues might impact the individual's savings. Hence, a functional dependency exists between the outstanding dues and the savings. So unless these relationships are taken into account, a counterfactual explanation of raising one's credit score might not result in being approved for the loan as in the process of doing so, one will affect the outstanding balance and savings. Even in the case that one does get approved for the loan, counterfactual explanations might require a higher amount of effort than required. For example, the loan approval system emphasizes outstanding dues and credit scores more than savings. In that case, clearing the dues to raise one credit score might put an individual in a situation where they have new savings (X$^{'}_{1}$) = \$20,000, outstanding dues(X$^{'}_{2}$) = $\$5,000$, and credit score(X$^{'}_{3}$) = 685 in which case for this new set of instances, the loan approval might only require a credit score of 690 versus 700. The example highlights the danger of not paying attention to the causal relationships that govern the world. 
\section{Counterfactual Explanations with s(CASP)}

% OLD
% 10/3
% SOPAM -Begin 1
% SOPAM -Mid 1
We illustrate our technique through the example of the well known adult income dataset \cite{adult}. Based on census data, it is a dataset for predicting whether a person {\tt P}'s income \texttt{`=<50k/yr'}. Given that a person's income is predicted to be \texttt{`=<50k/yr'} based on certain feature values, how can these feature values be changed so that {\tt P}'s income becomes \texttt{`>50k/yr'}? This type of counterfactual reasoning is what we are interested in.
%The problem that we are trying to solve is a special naive hypothetical case - \textit{Assuming that an individual wishes to ask for a pay raise. Given the data available and the decisions made by state-of-the-art machine learning models, what does one need to change in order to obtain the desired decision where the model says that you are someone who should earn `$>$\$50k/yr'?}.
%This is the hypothetical case where we assume that the company uses a classification model to decide what wage slab to put employees under and negotiate a salary.  

We first utilize the FOLD-SE algorithm \cite{foldse}, that learns a stratified answer set program to predict if a person's income will be \texttt{`=<50k/yr'}. 
Our objective is to obtain valid counterfactual explanations for any decision on an instance that satisfies these rules. That is, for an instance that obtains a prediction of \texttt{`=<50k/yr'}, we compute the counterfactual explanations that result in the model giving a different prediction, namely, \texttt{`>50k/yr'}. 

The adult dataset has 14 features that are a mixture of numerical and categorical values. The classification label indicates whether one makes \texttt{`=<50k/yr'}. We train the FOLD-SE rule-based algorithm to obtain two rules. 

\begin{lstlisting}[style=MySCASP, firstnumber=1,label=lst_FOLD-SE Rules, caption = {FOLD-SE Rules}]
label(X,'<=50K') :- not marital_status(X,'Married-civ-spouse'), 
        capital_gain(X,N1), N1=<6849.0.
label(X,'<=50K') :- marital_status(X,'Married-civ-spouse'), 
        capital_gain(X,N1), N1=<5013.0, education_num(X,N2), N2=<12.0.
\end{lstlisting}

% Original
\begin{comment}
\begin{lstlisting}[style=MySCASP, firstnumber=1]
less_equal_50K(X,Y,Z) :- marital_status(X), capital_gain(Y),
        X \= married-civ-spouse, Y #=< `6849.0`.
less_equal_50K(X,Y,Z) :- marital_status(X), capital_gain(Y),
       education_num(Z), X = married-civ-spouse, Y #=<`5013.0`, Z #=<`12.0`.
\end{lstlisting}  
\end{comment}

\medskip Line 1 says that a person whose marital status is not `Married-civ-spouse' with a capital\_gain less than or equal to \$6,849, will make less than \$50K per year (\texttt{`=<50k/yr'}). Line 3 says that a person with a marital status of `Married-civ-spouse,' with a capital\_gain of less than or equal to \$5,013 and with less than or equal to 12 years of education (education\_num) will make \texttt{`=<50k/yr'}. 

By utilizing the knowledge from the data and the FOLD-SE algorithm, we incorporate this knowledge into our s(CASP) system. Indisputable details and facts are represented as s(CASP) facts for, e.g., the domain of the features. The rest of the knowledge is represented as s(CASP) rules. Through the dual rules, the rules corresponding to negation of predicates in the rule heads are generated, which helps formulate the imagined alternate worlds.
Next, we describe our method for computing the counterfactuals in detail using the Adult dataset example.
% SOPAM -End 2

\subsection{Modeling Factual Knowledge}
% SOPAM -Begin 3

% SOPAM -Mid 3
Information from the training data is extracted and represented as facts. However, as shown in Listing \ref{lst_FOLD-SE Rules}, the decision depends only on three features. Thus, only the features mentioned in the rules should act as constraints in the decision-making process. The adult dataset has two types of features, shown below with their respective domains:
\begin{enumerate}

    \item Numerical Features:
    \begin{itemize}
        \item capital\_gain: [0, 99999]
        \item education\_num: [1, 16]
        \item age: [17, 90]
    \end{itemize}
    \item Categorical Features:
    \begin{itemize}
        \item marital\_status: [`Divorced', `Married-AF-spouse', `Married-civ-spouse', `Married-spouse-absent', `Never-married', `Separated', `Widowed']
        \item relationship: [`Husband', `Not-in-family', `Other-relative', `Own-child', `Unmarried', `Wife']
        \item sex: [`Male', `Female']
    \end{itemize}
\end{enumerate}

We represent the domain of the features as facts. For categorical features, we declare their domain using the {\tt{}f\_domain/2} predicate. For illustration, we shall only use a small subset of feature values in this paper, including those incorporated in the FOLD-SE rules.  
%ask Sopam 
\begin{lstlisting}[style=MySCASP, firstnumber=1]
f_domain(marital_status, married_civ_spouse).  
f_domain(marital_status, divorced).  

f_domain(relationship, husband).  
f_domain(relationship, wife).  
f_domain(relationship, unmarried).

f_domain(sex, male). 
f_domain(sex, female). 
\end{lstlisting}

%Original
\begin{comment}
\begin{lstlisting}[style=MySCASP, firstnumber=1]
% f_domain/2 determines the domain for every feature
%    Marital_status: 7 values (we only use 2 values for simplicity)
f_domain(marital_status, married_civ_spouse).  
f_domain(marital_status, divorced).  
    			
%    Relationship: 7 values (we only use 3)
f_domain(relationship, husband). 
f_domain(relationship, wife). 
f_domain(relationship, unmarried).
\end{lstlisting}  
\end{comment}

%f_domain(marital_status, never_married). %Never-married
    %f_domain(marital_status, separated). %Separated
    %f_domain(marital_status, widowed). %Widowed
    %f_domain(marital_status, married_spouse_absent). %Married-spouse-absent
    %f_domain(marital_status, married_AF-spouse). %Married-AF-spouse.
			
    %f_domain(marital_status, not_married_civ_spouse). %Married-civ-spouse

    %f_domain(relationship, own_child).
    %f_domain(relationship, not_in_family).
    %f_domain(relationship, other_relative). 
    %f_domain(relationship, not_husband_wife). %Divorced

\noindent Lines 1-2 define the domain of the categorical feature \code{marital_status}, Lines 4-6 and 8-9 define the same for categorical features \code{relationship} and \code{sex}.
%where the other four values are \code{own_child, not_in_family, other_relative, not_husband_wife}.
% SOPAM -End 3

\subsection{Modeling Categorical Features}

We have defined the domain for the categorical values. Our goal is to move from a world where a undesired outcome (\texttt{`=<50k/yr'}) was obtained to a world where we achieve a desired result (\texttt{`>50k/yr'}) through interventions. Hence, each feature has a value prior to intervention, and a value post-intervention. We explicitly represent the pre and post-interventions status of the features through two separate objects (before\_int and after\_int). However, at a given point of time, the value taken by a feature at any given moment is unique. This is captured via even loops over negation. For example, a person can only have a distinct \code{marital_status} value before intervention. If a person is divorced, he/she cannot be married as well. Similarly after intervention, a person can only have one value for \code{marital_status}. It might be the same value as before the intervention or it might change after the intervention. However, it cannot have multiple values at a given point of time. The exact property is incorporated in the pre-intervention world through modeled lines 2-4 and in the post-intervention world modeled through lines 7-9 below. They represent the value of \code{marital_status} feature in each of the worlds.

 %#show before_int_marital_status/1, not before_int_marital_status/1.
%    #show after_int_marital_status/1, not after_int_marital_status/1.

\begin{lstlisting}[style=MySCASP, firstnumber=1]
% Pre-intervention world
not_before_int_marital_status(X) :- f_domain(marital_status, Y),
        before_int_marital_status(Y), Y \= X. 
before_int_marital_status(X):- not not_before_int_marital_status(X).

% Post-Intervention world
not_after_int_marital_status(X) :- f_domain(marital_status, Y),
	after_int_marital_status(Y), Y \= X. 
after_int_marital_status(X):- not not_after_int_marital_status(X). 
\end{lstlisting} 

The above rules model the uniqueness of marital status in the pre- and post-intervention worlds. 
%\code{before_int_marital_status(X)} and \code{after_int_marital_status(X)}) define the pre-intervention world and the post-intervention world. 
In the pre-intervention world, one obtains the decision \texttt{`=<50k/yr'}, and in the post-intervention world, we may obtain the counterfactual explanation that gives the decision \texttt{`>50k/yr'}. 

\begin{comment}
Between the pre and post-intervention world, the features have specific properties:
\begin{itemize} 
    \item At a given time, only  a distinct value can be taken (mutual exclusivity at any given time).
    \item Across interventions, the values can stay the same or change. We put no constraint on the post-intervention value with respect to the pre-intervention value. 
\end{itemize}
\end{comment}
We similarly write the code for the categorical features relationship and sex.

% original
\begin{comment}
 %#show before_int_relationship/1, not before_int_relationship/1.
%    #show after_int_relationship/1, not after_int_relationship/1.
\begin{lstlisting}[style=MySCASP, firstnumber=42]
% relationship
% Pre-intervention world ie. 
before_int_relationship(X):- not not_before_int_relationship(X).
not_before_int_relationship(X):- f_domain(relationship, Y), 
         before_int_relationship(Y), Y \= X. 
    
% Post-intervention world 
after_int_relationship(X) :- not not_after_int_relationship(X).
not_after_int_relationship(X):- f_domain(relationship, Y), 
        after_int_relationship(Y), Y \= X.     
\end{lstlisting}   
\end{comment}

\subsection{Rules for Numerical Features}
Next we define the domains for numerical features. For domains of numerical features, we use builtin constraints of s(CASP) to specify the range of values of these features. 
%We do that directly through the numerical feature objects. The objects constrain the values taken by the features to a particular range. 
The range values for all the features were taken from the maximum and minimum values available from the dataset. Line 1 defines the range for the capital\_gain feature. Line 2 defines the range for education\_num and line 3 does the same for age.
%#show capital_gain/1, not capital_gain/1.
%#show education_num/1, not education_num/1.
%#show age/1, not age/1.

\begin{lstlisting}[style=MySCASP, firstnumber=1]
capital_gain(X):- X #>= 0, X #=< 99999.
education_num(X):- X #>= 1, X #=< 16.
age(X):- X #>= 17, X #=< 90.
\end{lstlisting}

\vspace{-0.1in}
\subsection{Modeling the decision making rules} 

We have run the FOLD-SE algorithm to obtain the rules corresponding to the label \texttt{`=<50k/yr'} as shown in Listing \ref{lst_FOLD-SE Rules}.

%Original

\begin{comment}
\begin{lstlisting}[style=MySCASP, firstnumber=1]
less_equal_50K(X,Y) :- marital_status(X), capital_gain(Y),
        X \= married-civ-spouse, Y #=< `6849.0`.
less_equal_50K(X,Y,Z) :- marital_status(X), capital_gain(Y),education_num(Z),
        X = married-civ-spouse, Y #=< `5013.0`, Z #=< `12.0`.
\end{lstlisting} 
\end{comment}

Both rules are mutually exclusive and contribute to the final decision. However, to make them compliant with our code, we rewrite the same rules as below:

\begin{lstlisting}[style=MySCASP, firstnumber=1, label= lst_lite_le_50K]
% Decision rule to classify if a person makes '<=50K/yr'
lite_le_50K(X,Y,_) :- X \= married_civ_spouse,  Y #=< `6849.0`.
lite_le_50K(X,Y,Z) :- X = married_civ_spouse, Y #=< `5013.0`, Z #=< `12.0`.		

less_equal_50K(A,B,C):- f_domain(marital_status, A), 
    before_int_marital_status(A), capital_gain(B), 
    education_num(C), lite_le_50K(A,B,C).
\end{lstlisting}

The new rule {\tt{lite\_le\_50K/3}} does not contain the feature objects as defined earlier but only the decision-making components (\code{\{X \= married_civ_spouse,  Y #=< 6849.0}\} and \code{\{X = married_civ_spouse,  Y #=< 5013.0, Z #=< 12.0}\}). These decision-making components and feature objects in conjunction help construct the various worlds. Hence, the worlds defined by the rule {\tt{less\_equal\_50K/3}} in line 5 will provide a decision where somebody will make \texttt{`=<50k/yr'}. Verification can be done by running the query. ?-\code{less_equal_50K(A,B,C)}.

%\ifflase
%GG 10/8: Causal relationships section can be removed as it is not needed for computing counterfactuals
\subsection{Modelling Causal Relationships}

%Begin Sopam 
We incorporate causal dependencies into our approach in the form of rules, and the resulting counterfactual explanation should model the downstream changes generated by the causal assumptions to give the desired counterfactual solutions that can be realized in reality.
% End Sopam
% 
%GG1012: Sopam, please check
\begin{comment}
We have run the FOLD-SE algorithm to obtain the feature dependency between certain features for the Adult dataset. We incorporate them into our program to obtain the rules capturing the dependency of `marital\_status' on `relationship' and `age'.
The causal rules are obtained by treating `marital\_status' as the target label (the original label for \texttt{`=<50k/yr'} is ignored), and the other attributes as feature, then running the FOLD-SE algorithm again with the new label.
   
\end{comment}

%% Sopam GG1012
Using the FOLD-SE algorithm, we obtain the following rules highlighting a dependence of the feature `marital\_status' on relationship' and `age.'
%% END SOPAM GG1012

\begin{lstlisting}[style=MySCASP]
% Constraint rules identify causal relationships amongst features.
% They restrict the values taken for relationship(Y) and age(Z)
constraint_ms_reln_age(married_civ_spouse,Y,Z):- Y = husband.
constraint_ms_reln_age(married_civ_spouse,Y,Z):- Y = wife.
constraint_ms_reln_age(never_married,Y,Z):- Y \= husband, Y\= wife, Z #=<29.
% Add the rule to catch all other cases
constraint_ms_reln_age(X,Y,Z):- X\= married_civ_spouse, X\= never_married, 
                                Y \= husband, Y\= wife.
\end{lstlisting}
% Original
\begin{comment}
\begin{lstlisting}[style=MySCASP]
% Corresp. constraint rules to identify causal relationships amongst features.
% Following rules that restrict the values taken for relationship(Y)	
constraint_ms_reln(X,Y):- X= married_civ_spouse,Y = husband.
constraint_ms_reln(X,Y):- X= married_civ_spouse,Y = wife.
% Add the rule to catch all other cases
constraint_ms_reln(X,Y):- X\= married_civ_spouse, Y \= husband, Y\= wife.
\end{lstlisting}
\end{comment}

% However since we have no rule for divorce or unmarried
% , etc, marital status will only produce married_civ_spouse.
% We add a rule to show that if one isn't married they aren't husbands or wives
Since we are limiting ourselves to only a restricted number of relationship values, we have to add a rule to catch all other situations, shown in line 7. Similarly, we learn the causal dependency that `relationship' has on `age' and `sex'.
% Original
\begin{comment}
\begin{lstlisting}[style=MySCASP, firstnumber=1]
% Corresponding constraint rules that restrict age for relationship (Y)	
% The rule constrains that husbands can ONLY be above 27 years of age.
constraint_reln_age(X,Y):- age(Y), X = husband, Y#>27.
% Since there is no rule for wife, we put a rule to allow the same
constraint_reln_age(X,Y):- age(Y), X \= husband.
\end{lstlisting}
\end{comment}

\begin{lstlisting}[style=MySCASP, firstnumber=1]
% Constraint rules that restrict age and sex for relationship
constraint_reln_sex_age(husband,Y,Z):- Y\= female, Z#>27.
% There is no rule for wife, so we put a rule to allow the same
constraint_reln_sex_age(wife,Y,Z):- Y= female.
% Add the rule to catch all other cases
constraint_reln_sex_age(X,Y,Z):- X \= husband, X \= wife.
\end{lstlisting}

\begin{comment}
\begin{lstlisting}[style=MySCASP, firstnumber=1]
% Corresponding constraint rules that restrict age and sex for relationship (Y)	
% The rule constrains that husbands cannot be Female and above 27 years of age.
constraint_reln_age(X,Y):- age(Y), X = husband, Y#>27.
% Since there is no rule for wife, we put a rule to allow the same
constraint_reln_age(X,Y):- age(Y), X \= husband.
\end{lstlisting}
   
\end{comment}

From our rules obtained, we learn that a husband cannot be a female greater than 27 years of age. We do not obtain any rule for any non-husband relationship. Hence, we add a new rule in line 4 and 6 that will catch the other cases.

\subsection{Modelling  counterfactual explanations}
We have decision making rules {\tt{lite\_le\_50K/3}} corresponding to the label \texttt{`=<50k/yr'} as defined in listing \ref{lst_lite_le_50K}. 
%
\begin{comment}
\begin{lstlisting}[style=MySCASP, firstnumber=1]
% Decision rule to classify if a person makes '<=50K/yr'
lite_le_50K(X,Y,_) :- X \= married_civ_spouse,  Y #=< `6849.0`.
lite_le_50K(X,Y,Z) :- X = married_civ_spouse, Y #=< `5013.0`, Z #=< `12.0`.		
\end{lstlisting}
\end{comment}
%
To obtain the counterfactual explanations, we utilize the dual rules automatically computed by s(CASP), which constructively compute a  value for the negated goal (\code{not lite_le_50K}). The dual rules are shown below.
\begin{lstlisting}[style=MySCASP, firstnumber=1]
not lite_le_50K(Var0,Var1,Var2) :-
   not o_lite_le_50K_1(Var0,Var1,Var2),not o_lite_le_50K_2(Var0,Var1,Var2).

not o_lite_le_50K_1(Var0,Var1,Var2) :- Var0 = married_civ_spouse.

not o_lite_le_50K_1(Var0,Var1,Var2) :- Var0 \= married_civ_spouse,
   Var1 #> `6849.0`.

not o_lite_le_50K_2(Var0,Var1,Var2) :- Var0 \= married_civ_spouse.

not o_lite_le_50K_2(Var0,Var1,Var2) :- Var0 = married_civ_spouse,
   Var1 #> `5013.0`.

not o_lite_le_50K_2(Var0,Var1,Var2) :- Var0 = married_civ_spouse,
   Var1 #=< `5013.0`, Var2 #> `12.0`.
\end{lstlisting}We obtain the counterfactual explanations by combining \code{not lite_le_50K} with the domain and feature objects.
\begin{lstlisting}[style=MySCASP, firstnumber=1]
% Counterfactual rule to clasify if a person does not make '<=50K/yr'
cf_less_equal_50K(A,B,C):- f_domain(marital_status, A),
            after_int_marital_status(A), capital_gain(B), 
            education_num(C), not lite_le_50K(A,B,C).
%% QUERY
?- cf_less_equal_50K(A,B,C).	
\end{lstlisting}

The new rule {\tt{cf\_less\_equal/3}}, is essentially identical to {\tt{less\_equal/3}} except that instead of {\tt{lite\_le\_50K/3}}, we use {\tt{not lite\_le\_50K/3}}. This translates to constructing an alternate world where the decision of somebody making \texttt{`=<50k/yr'} is not True. We use the object {\tt{after\_int\_marital\_status/3}} instead of {\tt{before\_int\_marital\_status/3}} to indicate that this is the post-intervention world.
We can verify the counterfactual explanations by invoking the query in lines~6, \code{?- cf_less_equal_50K(A,B,C)}.

\subsection{Modeling the intervention constraints} 

Assuming that an individual (instance) is predicted to make \texttt{`=<50k/yr'}(undesired decision), we aim to map the original instance with an undesired outcome to the counterfactual instance with the desired outcome (\texttt{`>50k/yr'}). To do that, we define the control rules that map the change across worlds (before/pre-intervention to after/post-intervention). 

The control variables have the property of making features either actionable or immutable across interventions. Using this flexibility, we assign a value of 0 to constrain a feature to be immutable across an intervention. If we do not assign a value to the variables, they shall be actionable. To enforce change on features after the intervention, i.e., to intervene on a feature resulting in a change in its value, we assign a value of 1 (categorical) or 1/-1 (numerical).
\begin{comment}
\begin{lstlisting}[style=MySCASP, firstnumber=1]
% Control Features: 
%      0: Features Remain Unchanged
%      1: Features will be changed
%      Category: 1- New CF feauture WILL take a different value

f_domain(control,0).
f_domain(control,1).
	
compare_C(X,X,0).
compare_C(X,Y,1):- X \= Y.
\end{lstlisting} 
\end{comment}
\begin{lstlisting}[style=MySCASP, firstnumber=1]
% Control Features: 
compare_C(X,X,0).
compare_C(X,Y,1):- X \= Y.
\end{lstlisting}

With numerical features, it is similar to that of the categorical case. However, there is a slight difference; we assign a value of +1 if the intervened feature should only increase, -1 if it should only decrease, and 0 to remain unchanged across interventions.
% Control Features: 
% 0: Features Remain Unchanged
% Numbers: +1- New CF feature will be greater than original
% Numbers: -1- New CF feature will be less than original
\begin{lstlisting}[style=MySCASP, firstnumber=1]
compare_N(X,X,0).
compare_N(X,Y,1):- X #< Y.
compare_N(X,Y,-1):- X #> Y.
\end{lstlisting}

These control rules provide a mapping between the pre-intervention and post-intervention world. They can also compute the Craig interpolant, highlighting what features must be intervened on to resolve the inconsistency with the decision-making rules. This is done by finding the minimum inconsistency between an instance and the counterfactual world. To find this we define a function {\tt{measure/7}} to measure the cost of making interventions:

\begin{lstlisting}[style=MySCASP, firstnumber=1]
% Measure
f_domain(control,0).
f_domain(control,1).
f_domain(control_N,0).
f_domain(control_N,1).
f_domain(control_N,-1).
measure(Z1,Z2,Z3,Z4,Z5,Z6,X):- f_domain(control,Z1), f_domain(control_N,Z2), 
    f_domain(control_N,Z3), f_domain(control,Z4),f_domain(control,Z5),
    f_domain(control_N,Z6), Q2 #= Z2*Z2, Q3 #= Z3*Z3,Q6 #= Z6*Z6,
    X #= Z1+Q2+Q3+Z4+Z5+Q6.
\end{lstlisting}

The rule {\tt{measure/7}}, takes the control variables(Z1-Z6) and computes the cost of intervention(X). We ideally want as low a cost as possible. For an instance that originally obtained an undesired outcome (\texttt{`=<50k/yr'}), the counterfactual explanations corresponding to the lowest cost denote the proof of inconsistency which upon resolving takes a factual instance to the counterfactual space. This is the Craig Interpolant for the factual instance and the counterfactual space. 

%% To do now 10/4
\subsection{Modelling Rules for the Decision and the Counterfactual}
To obtain  the counterfactual instances for a given decision, we first need to obtain the original decisions, i.e., decisions corresponding to the classification that a person makes \texttt{`=<50k/yr'}. To do that, we use the decision rules we defined in listing \ref{lst_FOLD-SE Rules}

%Original
\begin{comment}
   \begin{lstlisting}[style=MySCASP, firstnumber=1, basewidth=.45em]
% Removing NAF from working_class/3
% 3 features matiral status and capital gain
lite_working_class(X,Y,_) :- X \= married_civ_spouse,  Y #=< `6849.0`.
lite_working_class(X,Y,Z) :- X = married_civ_spouse, Y #=< `5013.0`, Z #=< `12.0`.		
\end{lstlisting} 
\end{comment}
We have shown how to obtain decisions as well as counterfactual explanations. We aim to highlight a mapping between our pre-intervention world (where we are subject to an unwanted decision) and our post-intervention world (where we obtain the desired outcome). We write the following rule to highlight what features need to be intervened on to get our desired outcome.

%    # show lite_le_50K/3, not lite_le_50K/3.

\begin{lstlisting}[style=MySCASP, firstnumber=1]
craig(A,B,C,D,E,F,Z1,Z2,Z3,Z4,Z5,Z6,A1,B1,C1,D1,E1,F1,X):-  
        f_domain(marital_status, A),f_domain(relationship, D), 
        f_domain(sex,E), before_int_marital_status(A), 
        capital_gain(B), education_num(C), 
        before_int_relationship(D), before_int_sex(E), age(F), 
        constraint_ms_reln_age(A,D,F), constraint_reln_sex_age(D,E,F),     
		f_domain(marital_status, A1),f_domain(relationship, D1), 
        f_domain(sex,E1), after_int_marital_status(A1), 
        capital_gain(B1), education_num(C1), 
        after_int_relationship(D1), after_int_sex(E1), age(F1) , 
        constraint_ms_reln_age(A1,D1,F1), 
        constraint_reln_sex_age(D1,E1,F1),
		compare_C(A,A1,Z1), compare_N(B,B1,Z2), compare_N(C,C1,Z3), 
        compare_C(D,D1,Z4), compare_C(E,E1,Z5), compare_N(F,F1,Z6),
		measure(Z1,Z2,Z3,Z4,Z5,Z6,X), 
        lite_le_50K(A,B,C), not lite_le_50K(A1,B1,C1).
\end{lstlisting}

The above rule {\tt{craig/19}} for the given input features marital\_status, capital\_gain, education\_num, relationship, sex and age represented by argument A, B, C, D ,E and F respectively provides the counterfactual instance characterized by the features A1, B1, C1, D1, E1 and F1. The arguments Z1, Z2, Z3, Z4, Z5 and Z6 indicate whether features A, B, C, D, E and F have been intervened on. The argument X denotes the cost of intervention. X tracks the count of the features intervened on. By definition, X has to be greater than 0 as the pre and post-intervention worlds are inconsistent. For this scenario, the maximum value for X is 6 (as there are six features to be intervened on). In order to obtain a suggestion on what to change, we run the query \code{?-craig(A,B,C,D,E,F,Z1,Z2,Z3,Z4,Z5,Z6,A1,B1,C1,D1,E1,F1,X)}.

\section{Performance Evaluation}
% SOPAM -Begin 4
A counterfactual explanation is possible if the query returns a model. We obtain various (partial) answer sets from the knowledge, rules, and dual rules encoded in our system, as well as information on how to achieve the desired counterfactual explanation and its cost. We see the counterfactual explanation for an instance that originally obtained an undesired outcome ($=<$\$50K/yr'). In addition, we can also compute Craig's Interpolant to find the inconsistency that needs to be resolved to achieve the desired outcome (to achieve the counterfactual solution by traversing to the counterfactual world).
% SOPAM -Mid 4

\begin{figure}[htp]
    \centering
    \includegraphics[width=12cm]{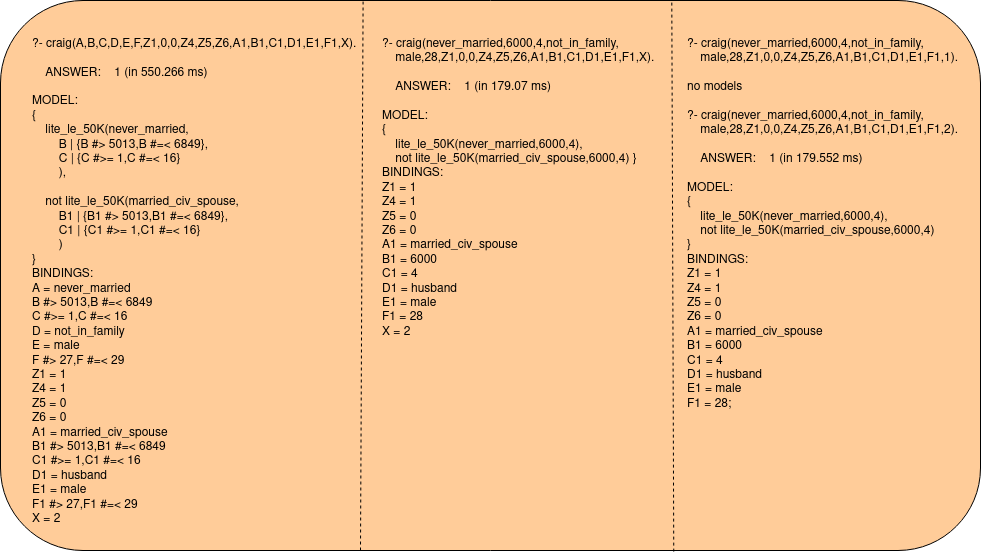}
    \caption{\textbf{Left:} All possible transitions from \texttt{`=<50k/yr'} to \texttt{`>50k/yr'}; \textbf{Middle:} CFE for given individual and; \textbf{Right:} Craig's Interpolant: CFE with minimum cost}
    \label{rslt_query}
\end{figure}

\medskip\noindent\textbf{Obtaining all possible counterfactual explanations:}
To obtain all possible counterfactual explanations for individuals that make \texttt{`=<50k/yr'}, we query {\tt{craig/19}} as shown in the left column of Fig. \ref{rslt_query}. The resulting counterfactual solutions, where all parameters are unbound, highlight all possible factual values for the instances (A,B,C,D,E,F), the counterfactual instances (A1,B1,C1,D1,E1,F1), the control values (Z1,Z2,Z3,Z4,Z5,Z6) that indicate whether a feature is intervened on, and finally the cost of intervention (X).

\vspace{-0.2in}

\noindent\textbf{Obtaining the counterfactual explanations for a given individual:}
In order to obtain the counterfactual explanations for an individual that makes \texttt{`=<50k/yr'}, we query {\tt{craig/19}} as shown in the middle column of figure \ref{rslt_query}. 
An individual with attributes marital\_status = never\_married, capital\_gain = 6000, education\_num = 4, relationship = not\_in\_family, gender = male and age = 28 is classified as someone earning \texttt{`=<50k/yr'} (not desired by the individual). This is shown through the proof for {\tt{lite\_le\_50K/3}}
If this individual is flexible about changing everything except for the capital\_gain and education\_num, we can see if there does indeed exist a counterfactual explanation as shown.

%\begin{figure}[htp]
%    \centering
%    \includegraphics[width=9cm]{pictures/Craig_Suggestion.png}
%    \caption{Obtain counterfactual explanations for an individual given his preference}
%    \label{rslt_craig_suggestion}
%\end{figure}

\medskip\noindent\textbf{Obtaining Craig's Interpolant:}
In order to obtain Craig's Interpolant, we aim to get a proof for the inconsistency. The proof highlights the minimum inconsistency that needs to be resolved. We can find the minimum inconsistency by checking for the minimum cost $X$ that gives us a counterfactual explanation. This is shown in the right column in figure \ref{rslt_query}. Since by default the counterfactual explanation and the factual explanation belong to worlds that are inconsistent, the minimum cost cannot be 0. The maximum cost corresponds to all the features being intervened on (6). So we run the query for the individual with a cost of $X=1$ to see if such a counterfactual explanation exists. The query fails to return a model, so we run the same query but for $X=2$. The resulting counterfactual explanation corresponds to Craig's Interpolant highlighted by the control variables (Z1,Z2,Z3,Z4,Z5,Z6) which highlight the features in which the inconsistencies lie.

%\begin{figure}[htp]
%    \centering
%    \includegraphics[width=9cm]{pictures/Craig_Interpolant.png}
%    \caption{The minimum cost $X$ for interventions denotes the Craig's Interpolant}
%    \label{rslt_craig_interpolant}
%\end{figure}

%\code{?- craig(A,B,C,D,E,Z1,Z2,Z3,Z4,Z5,A1,B1,C1,D1,E1,X)}.

    %\begin{figure}[htp]
    %    \centering
    %    \includegraphics[width=12cm]{pictures/Paddle_fig1.png}
    %    \todo[inline]{Use lstlisting env. for the results --is nicer --copy-paste the output from the console below ;-)}
    %    \begin{lstlisting}[style=Plain, firstnumber=1]
    %% QUERY:?- craig(A,B).

    %        ANSWER: 1 (in ...)

    %MODEL:
    %{ before_int_marital_status(married_civ_spouse), capital_gain(B | {B #>= 0, B #=< 5013}) }
    %BINDINGS:
    %A = married_civ_spouse
    %B #> 0,B #=< 5013
    %C #> 1,C #=< 12 ? ;
    %\end{lstlisting}
    %    \caption{Justification for a Counterfactual through what rules are proved}
    %    \label{fig:galaxy}
    %\end{figure}

%\begin{lstlisting}[style=Plain, firstnumber=1]
%% QUERY:?- craig(A,B,C,D,E,Z1,Z2,Z3,Z4,Z5,A1,B1,C1,D1,E1).

%        ANSWER: 1 (in 108.983 ms)

%MODEL:
%{}
%BINDINGS:
%A = married_civ_spouse
%B #> 0,B #=< 5013
%C #> 1,C #=< 12 
%D = husband
% #>27, E #=< 90
%Z1 = 0
%Z2 = 0
%Z3 = 1
%Z4 = 0
%Z5 = 0
%A1 = married-civ-spouse
%B1 #>= 0, B1 =< 5013
%C1 #> 12, C1 =< 16
%D1 = husband
%E1 #> 27, E1 =< 90
%X = 1 ? ;
%\end{lstlisting}

% Sopam Begin

\vspace{-0.1in}

\section{Experimental Evaluations}

We have tested our counterfactual explanation algorithm on $6$ datasets. We check the performance of our algorithm in generating the counterfactual explanations as the domain size of the features increase. In addition to this we also view the increase in complexity when considering additional features for the computation of counterfactual explanations such as when we take causal relations into account. 
%This is highlighted by taking the non-causal case 
%as the number of features increases as well as when the number of different values that a feature can take increases. 
As the number of features and the domain size of features rise, so do the number of possible world. For example, when a binary feature's domain size increases to three, the search space increases. This makes sense as when the features have greater flexibility the complexity in computing the appropriate counterfactual explanation increases. This is analogous to how we as humans think. If the possible options available increases, so does the complexity in deciding the best course of action. 

We run the experiments on the following datasets: Adult \cite{adult}, Titanic \cite{titanic}, Car Evaluation \cite{car}, Academic dropout \cite{dropout}, Voting \cite{voting} and Mushroom \cite{mushroom}. We measure the average time it takes to produce a counterfactual explanation/binding as the number of feature values increase.

\subsection{Time versus Feature Values}
By using \textit{even-loops-over-negation (ELON)}, we are able to generate multiple worlds. As the number of feature values increase, so does the time taken to compute the counterfactual explanations.  We verify this through running the experiments to find the counterfactual explanations for various datasets. 
%\vspace{-0.1in}
\begin{table}[h!]
  \centering
  \resizebox{\columnwidth}{!}{%
  \renewcommand{\arraystretch}{2}
  \begin{tabular}{|p{2cm}|c|c|c|c|c|c|c|c|c|c|c|c|c|c|c|c|c|c|c|}
    \hline
    \multicolumn{1}{|c|}{\textbf{Dataset}} & \multicolumn{4}{c|}{Adult} & \multicolumn{3}{c|}{Mushroom}   & \multicolumn{3}{c|}{Cars}  & \multicolumn{3}{c|}{Titanic}  & \multicolumn{3}{c|}{Voting} & \multicolumn{3}{c|}{Dropout} \\
    \cline{1-20}
    \hline
    \multicolumn{1}{|c|}{\textbf{Feature}} & \multicolumn{4}{c|}{marital\_status}  & \multicolumn{3}{c|}{odor} & \multicolumn{3}{c|}{maint}   & \multicolumn{3}{c|}{pclass}  & \multicolumn{3}{c|}{education\_spending}  & \multicolumn{3}{c|}{admission\_order} \\
    \cline{1-20}
    \multicolumn{1}{|c|}{\textbf{Domain Size}} & 2 & 3 & 4 & 5 & 2 & 3 & 4 & 2 & 3 & 4& 1 & 2 & 3& 1 & 2 & 3& 2 & 3 & 4\\
    \hline
    \multicolumn{1}{|c|}{\textbf{Avg. Time(ms)}} & 175  & 201 & 238 & 280 & 727 & 952 & 1392 & 524 & 620 & 709 & 76 & 104 & 165 & 470 & 559 & 841 & 698 & 899 & 1145 \\ \hline

  \end{tabular} 
  }
  \vspace{0.05in}
  \caption{Time vs. Domain Size: Adult,Mushroom,Cars,Titanic,Voting,Dropout}
  \label{tbl_time_feature_values}
  \vspace{-0.2in}
\end{table}

For the Titanic dataset \cite{titanic}, we keep all the features and their domain size constant (categorical) except for the feature \textbf{`pclass'.} We obtain the solution for the interventions to generate the counterfactual explanatins for the cases where \textbf{`pclass'} initially has a domain size of 1 and keep increasing the the domain size. We measure the average time taken to compute a counterfactual explanation. We do the same for the Voting, Cars, Dropout, Adult and Mushroom datasets. As seen in Table \ref{tbl_time_feature_values}, when the domain size of features increase, so does the complexity in computing the counterfactual explanation and hence the time taken.

\subsection{Consequence of Causality: Additional feature complexity}

As mentioned in the previous section, \textit{even-loops-over-negation (ELON)} facilitates the creation of multiple worlds. As the number of features increases, so do the possible worlds. For example, if the decision to go out depends only on our health, we imagine two worlds, one where we are healthy and one where we are not. However, if the decision relies on two factors, health and the weather (sunny or rainy), then we imagine $2^2$ worlds. Hence, the number of worlds imagined and the computation time increases. We verify this on the Adult dataset. Specifically, three features are used in the non-causal case where the categorical feature marital\_status contributes to generating multiple worlds. Feature independence is assumed; hence, only decision-making features generate counterfactual explanations. When we take causal relations into account, the features that causally influence the decision-making are taken in the computation of the counterfactual explanation. In the causal case, six features are used where the categorical features marital\_status, relationship, and sex contribute to generating possible worlds. We compare the performance in generating counterfactual explanations for the two cases. When considering causal relations, a more extensive set of features (6) is used, where three are categorical, and the non-causal case uses only the decision-making features (3), out of which one is categorical. From Table \ref{tbl_adult_1_3}, we see that when considering causal relations, the influence of the additional features due to feature dependency increases the complexity of finding the counterfactual explanations as per our expectations.

\vspace{-0.15in}
\begin{table*}[h!]
\centering
% \fontsize{9}{10}\selectfont
\setlength{\tabcolsep}{6.0pt}
\begin{tabular}{@{}rlcr@{}}
\toprule
\multicolumn{1}{l}{Dataset} & Categorical features used & count  & Time (ms) \\ \midrule

\multirow{1}{*}{Adult}      & marital\_status & $1$  & $281$ \\ %\cmidrule(lr){1-4}                        

\multirow{1}{*}{Adult}      & marital\_status, relationship, sex & $3$ & $709$ \\ 
\bottomrule                        
\end{tabular}
\vspace{.5em}
\caption{Non-Causal vs. Causal Case}
\label{tbl_adult_1_3}
\label{tb_1}
\end{table*}

\vspace{-0.5in}

\section{Related Work}
% New
% 10/4 Post Grammarly
There have been existing approaches to tackle this problem of a lack of transparency by providing an explanation to an undesired outcome such as that of Wachter et al. \cite{wachter}. Some approaches are tied to particular models or families of models, while some use optimization-based approaches to try and solve the same problem \cite{ref_mace_1,ref_mace_2}. Ustun et al \cite{recourse_1} took an approach that highlighted the need to focus on algorithmic recourse  to ensure a viable counterfactual explanation. Others have found ways to utilize counterfactual explanations to improve model performance and ensure accurate explanations such as White and Garcez \cite{ref_clear}. Lately, specific approaches have considered the context of consequential decision-making concerning the type of features being altered and focused on producing viable realistic counterfactuals such as the work by Karimi et al. \cite{ref_mace}. Our approach utilizes abductive reasoning to provide possible explanations that justify a decision through counterfactual explanations by utilizing the answer set programming paradigm (ASP). While there have been other approaches utilizing ASP such as Bertossi and Reyez \cite{ref_asp_cf}, it does not make use of a goal directed ASP system and relies on grounding which has a disadvantage of losing the association between variables. Our approach utilizes s(CASP) and hence no grounding is required. Additionally we can implement complex logic over the control variables (Z1-Z6). A special case of this obtains the Craig interpolant for a failing query. 
%Note that there are other proposals for goal-directed implementation of ASP \cite{goaldir1,goaldir2,goaldir3}. 
% GG Said not to use predicate and propositional as some it might leave a bad tadste in some purists' mouth- However, they can only handle propositional answer set programs. The s(CASP) system \cite{sCASP_org} can execute predicate answer set programs without grounding. 
The main contribution of this paper is to show that complex tasks such as imagining possible scenarios, which is essential in generating counterfactuals as shown by Byrne \cite{ref_Byrne_CF} (``what if something else happened?"), can be modeled with s(CASP). Such imaginary worlds (where alternate facts are true and, hence, different decisions are made) can be automatically computed using the s(CASP) system. 
%In addition to commonsense reasoning applications \cite{ref_GG} of the most complex type, this can be modeled using s(CASP) \cite{bim-arias22,arias-ec2022}. 
Detailed explanations are also provided by s(CASP)  for any given decision reached. The s(CASP) system thus allows for counterfactual situations to be imagined and reasoned. We assume the reader is familiar with ASP and s(CASP). An introduction to ASP can be found in Gelfond and Kahl's book \cite{gelfond-kahl}, while a fairly detailed overview of the s(CASP) system can be found elsewhere \cite{arias-ec2022,scasp-iclp2018}.

\section{Conclusion}

In this paper, we addressed the problem of automatically generating counterfactual explanations. Given that more and more decisions are being made with the help of models learned using machine learning systems, the problem of interpretability and explainability is becoming increasingly important. Another major problem related to explainability is the problem of automatically determining changes that must be made to the input feature values in order to flip the decision from a negative one to a positive one. Computing these changes amounts to finding counterfactual explanations. We have shown how ASP and specifically the s(CASP) goal-directed ASP system can be used for generating these counterfactual explanations. The s(CASP) system's support for negation as failure and dual rules allow us to generate alternate worlds via abductive reasoning. To generate counterfactual explanations, we need to find alternate worlds that are reachable from the current world. We showed how we can use Craig Interpolation theorem to compute suggestions in the form of interventions to obtain a counterfactual explanation, if it exists. In addition this method can be used to find the the Craig Interpolant of a failing query. 
%By finding the minimum cost of intervention for generating a counterfactual, the control rules in our query highlight the minimum features to intervene on to resolve the inconsistency. 
\bibliographystyle{splncs04}
\bibliography{references,clip,general}

\end{document}